\documentclass[review]{elsarticle}
\usepackage{caption}
\usepackage{lineno,hyperref}
\usepackage{amsmath}
\usepackage{amssymb}
\usepackage{textcomp}
\usepackage{hyperref}
\usepackage{url}
\usepackage{graphicx}
\usepackage{rotating}
\usepackage{lscape}
\usepackage{longtable}
\usepackage{pdflscape}

\modulolinenumbers[5]
\makeatletter
\newcommand*{\centerfloat}{%
 \parindent \z@
 \leftskip \z@ \@plus 1fil \@minus \textwidth
 \rightskip\leftskip
 \parfillskip \z@skip}
\makeatother

\journal{Journal of Neuroscience Methods}

\begin{document}

\begin{frontmatter}

\title{Locomotion and gesture tracking in mice and small animals for neurosceince applications: A survey}

\author[uoc]{W. Abbas\corref{mycorrespondingauthor}}
\author[uoc]{D. Masip}
\ead{abbas@uoc.edu}

\address[uoc]{Department of Computer Science, Universitat Oberta de Catalunya, Barcelona, Spain.}

\begin{abstract}
Neuroscience has traditionally relied on manually observing lab animals in controlled environments. Researchers usually record animals behaving in free or restrained manner and then annotate the data manually. The manual annotation is not desirable for three reasons; one, it is time consuming, two, it is prone to human errors and three, no two human annotators will 100\% agree on annotation, so it is not reproducible. Consequently, automated annotation of such data has gained traction because it is efficient and replicable. Usually, the automatic annotation of neuroscience data relies on computer vision and machine leaning techniques. In this article, we have covered most of the approaches taken by researchers for locomotion and gesture tracking of lab animals. We have divided these papers in categories based upon the hardware they use and the software approach they take. We also have summarized their strengths and weaknesses.

\end{abstract}


\end{frontmatter}

\section{Introduction}
Neurosceince has found an unusual ally in the form of computer science which has strengthened and widened its scope. The wide availability and easy-to-use nature of video equipment has enabled neuroscientists to record large scale behavioral data of animals and analyze it from neurosciecne perspective. Traditionally, neuroscientists would record videos of animals which they wanted to study and then manually annotate the video data themselves. Normally this approach is reasonable if the video data being annotated is not large, but it becomes very inconvenient, tiresome, erroneous and slow as the amount of data increases. This is mainly because the annotations made by human annotators are not perfectly reproducible. Two annotations of the same sample done by two different persons will likely differ. Even the annotation done for the same sample at different times by same person might not be exactly the same. All of these factors have contributed to the demand of a general purpose automated annotation approach for video data. For behavioral phenotyping and neuroscience applications, researchers are usually interested in gesture and locomotion tracking. Fortunately, computer science has answers to this problem in the form of machine learning and computer vision based tracking methods. The research in this area is still not mature, but it is receiving a lot of attention lately. Primary motivation for automated annotation is the reproducibility and ability to annotate huge amounts of data in practical amount of time. 
\par The field is not mature. There is no consensus on which approach to follow yet, but most of the researchers follow  a loose set of rules. Some researchers approach this problem by treating video as sequence of still images and then applying computer vision algorithms to every frame in succession without considering their temporal relationship. Some of the researchers include temporal information to some extent while some approach towards it with the assistance of additional hardware. The general framework is similar. Animals (mice/rats/insects) are kept in a controlled environment, either restrained or free where the lighting and illumination can be manipulated. In order to acquire the video data, single or multiple video cameras are installed. These might be simple video cameras or depth cameras. There might be some additional accessories such as physical markers or body mounted sensors.
\label{sec:introduction}

\section{Problem Statement}
\begin{figure} 
  \centering
    \includegraphics[width=0.5\textwidth]{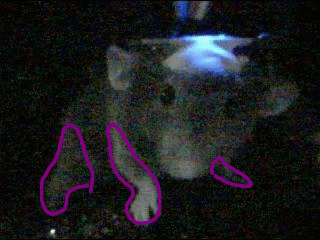}
    \caption{Frontal view of a mouse with its moving limbs marked}
    \label{FrontalSample}
\end{figure}

\begin{figure} 
  \centering
    \includegraphics[width=0.5\textwidth]{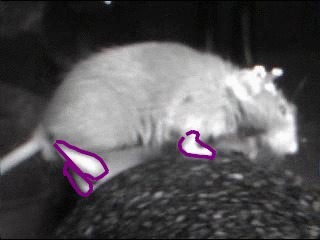}
    \caption{Lateral view of a mouse with its moving limbs marked}
    \label{LateralSample}
\end{figure}

Behavioral phenotyping depends upon annotated activity data of rodents. We can identify the activity type of a mouse when we see how it moves, behaves and acts over an extended period of time. One of many proposed approaches is to track the limb movements of the rodents and convert them into quantifiable patterns. The limbs tracking can be either achieved by recording them from frontal, lateral, top or bottom view. Typical tracking example from frontal and lateral view is shown in Fig. \ref{FrontalSample} and \ref{LateralSample}.

Cases shown in \ref{FrontalSample} and \ref{LateralSample} are typical examples of activity tracking in rodents and small animals. They present the following challenges
\begin{enumerate}
\item Spatial resolution in most consumer grade video cameras which have enough temporal resolution is not enough for effective tracking
\item The limbs might move faster at one point in time while they might be stationary at another point in time, rendering the development of uniform motion model impossible.
\item The limbs might overlap with each other or other body parts, therefore presenting occlusions.
\end{enumerate}

\section{Motion tracking principles in videos}
Videos are sequences of images/frames which if displayed with high enough frequency will be perceived as continuous content by the human eye. Although the video content appears continuous, it is still comprised of discrete images to which all the image processing techniques can be applied. Besides, the contents of two consecutive frames are usually closely related. The fact that video frames are closely related in spatial and temporal domains makes object and motion tracking possible in videos. Motion/Object tracking in video started with detecting objects in individual frames which in turn can be used for object tracking in video sequences. It involves monitoring an object's shape and motion path in every frame. This is achieved by solving the temporal correspondence problem, to match region in successive frames of a video sequence.
\par Motion detection is very significant when it comes to object tracking in video sequences. One one hand, motion adds another dimension to already complex problem of object detection in the form of object's temporal change requirements, on the other hand, it also provides additional information for detection and tracking. There are numerous researchers actively working on this problem with different approaches. Most of these methods involve single or multiple techniques for motion detection. They can be broadly classified three major categories; background subtraction and temporal differencing based approaches, statistical approaches and optical flow based approaches.

\subsection{Background Subtraction and Temporal Differencing}
Commonly used for motion segmentation in static scenes, background subtraction attempts to detect and track motion by subtracting the current image pixel-by-pixel from a reference/background image. The pixels which yield difference above a threshold are considered as foreground. The creation of the background image is known as background modeling. Once the foreground pixels are classified, some morphological post processing is done to enhance the detected motion regions. Different techniques for background modeling, subtraction and post processing results in different approaches for the background subtraction method \citep{BkgPaper1,BkgPaper2,BkgPaper3,BkgPaper4,BkgPaper5}.
\par In temporal differencing, motion is detected by taking pixel-by-pixel difference of consecutive frames (two or three). It is different from background subtraction in the sense that the background or reference image is not stationary. It is the mainly used in scenarios involving a moving camera \citep{BkgPaper6,BkgPaper7,BkgPaper8,BkgPaper9,BkgPaper10}.

\subsection{Statistical approaches}
Statistical methods are inspired by background subtraction methods in terms of keeping and updating statistics of the foreground and background pixels. Foreground and background pixels are differentiated by comparing pixel statistics with that of background model. This approach is stable in the presence of noise, illumination changes and shadows \cite{StatPaper1,StatPaper2,StatPaper3,StatPaper4,StatPaper5,StatPaper6,StatPaper7,StatPaper8,StatPaper9,StatPaper10}. 

\subsection{Optical Flow}
Optical  flow is the distribution of apparent velocities of
movement of brightness patterns in an image. It can arise from relative motion of objects and the observer. Consequently, it can give spatial and temporal information about various objects in the video \cite{FlowPaper1,FlowPaper2}. Optical flow methods exploit the flow fields of moving objects for motion detection. In this approach, the apparent velocity and direction of every pixel is be computed \cite{FlowPaper3,FlowPaper4,FlowPaper5,FlowPaper6,FlowPaper7}. Optical flow based methods can detect motion in video sequences when the observer is stationary or moving, however, most of the optical flow methods are computationally complex and cannot be used in real-time without specialized
hardware.

\section{Major trends}
Motion tracking for neuroscience applications can be treated as a special case of motion tracking; which means that all the motion tracking techniques can be applied to it in one way or the other. Although the general idea is the same, the environment for such type of motion tracking can be different. A typical setup includes a closed environment (either a room or a box), video cameras, the animal and control systems. The animal can either be restrained or freely behaving. There might be just a single camera or multiple cameras which records the motion from different angles. For this survey, we will go through all those cases which involves motion tracking (especially limbs tracking, head tracking and gesture tracking) of laboratory animals for behavioral phenotyping or medical assessment purposes. Based on their intended use and nature, we have divided the approaches found in literature in following categories.
\begin{enumerate}
\item Commercially available solutions
\item Hardware based methods
\item Video tracking aided by hardware
	\begin{enumerate}
	\item Semi-automated
    \item Completely automated
	\end{enumerate}
\item Video tracking methods mostly dependent on software based tracking
	\begin{enumerate}
	\item Semi-automated (aided by users or markers)
    \item Completely automated
	\end{enumerate}
\end{enumerate}

\section{Commercially available solutions}
\par We have covered commercially available solutions or approaches in this section. These solution includes all those hardware and software based methods which are available on demand from specific companies. Noldus corporation (\url{http://www.noldus.com/}) and CleverSyns (\url{http://cleversysinc.com/CleverSysInc/}) are two of the prominent names involved in the development of behavioral research technologies. The Orthotic Group (\url{http://www.theorthoticgroup.com/}) is involved primarily in gait analysis of humans but their approaches are exportable to gait analysis in rodents and small animals too. The Mouse Specific Inc. (\url{https://mousespecifics.com/}) deals primarily with behavioral research technologies for rodents. A few of the commercially available solutions are summarized based on the white papers from their parent companies.
\par Dorman et. al. did a comparison of two hardware assisted gait analysis systems; DigiGait and TreadScan \cite{PaperDigiGaitAndTreadScan}. The $DigiGait^{TM}$ imaging system uses a high-speed, 147 frames-per-second video camera mounted inside a stainless steel treadmill chassis below a transparent treadmill belt to capture ventral images of the subject. The treadmill is lit from the inside of the chassis by two fluorescent lights and from overhead by one fluorescent light. The $TreadScan^{TM}$ imaging system uses a high-speed, 100 frames-per-second video camera adjacent to a translucent treadmill belt to capture video reflected from a mirror mounted under the belt at $45^0$. Images are automatically digitized by $DigiGait^{TM}$ and $TreadScan^{TM}$ systems. $DigiGait^{TM}$ videos are manually cropped and imported then automatically analyzed. The software identifies the portions of the paw that are in contact with the treadmill belt in the stance phase of stride as well as tracks the foot through the swing phase of stride. Measures are calculated for 41 postural and kinematic metrics of gait. The authors found that $DigiGait^{TM}$ system consistently measured significantly longer stride measures $TreadScan^{TM}$. Both systems’ measures of variability were equal. Reproducibility was inconsistent on both systems. Only the $TreadScan^{TM}$ detected normalization of gait measures and the time spent on analysis was dependent on operator experience. $DigiGait^{TM}$ and $TreadScan^{TM}$ has been particularly well received in neuro-physiological research \cite{PaperDigiGait1,PaperDigiGait2,PaperDigiGait3,PaperDigiGait4,PaperDigiGait5,PaperDigiGait6,PaperDigiGait7} and \cite{PaperTreadScan1,PaperTreadScan2,PaperTreadScan3,PaperTreadScan4,PaperTreadScan5}.

\par Cleversys Inc. introduced a commercial solution for gait analysis in rodents, called GaitScan \cite{PaperGaitScan}. GaitScan system records video of the rodent running either on a transparent belt treadmill or on a clear free‐walk runway. The video of the ventral (underside) view of the animal is obtained using a high‐speed digital camera. The video essentially captures the foot prints of the animal as they walk/run. GaitScan software can work with videos taken from any treadmill or runway device that allows the capture of its footprints on any video capturing hardware system with a high‐speed camera. The accompanying software let the user track multiple gait parameters which can be later used for behavioral phenotyping. This solution has also been used in multiple researches \cite{PaperGaitScan1,PaperGaitScan2,PaperGaitScan3,PaperGaitScan4,PaperGaitScan5}.

\par TrackSys ltd. introduced two systems for rodents motor analysis. One system is called 'ErasmusLadder'. The mouse traverses a horizontal ladder between two goal boxes. Each rung of the ladder contains a touch-sensitive sensor. These sensors allow the system to measure numerous parameters relative to motor performance and learning such as step time and length, missteps,back steps and jumps \cite{PaperELadder}. It has been used in multiple researches \cite{PaperELadder1,PaperELadder2,PaperELadder3,PaperELadder4,PaperELadder5}. Its tracking performance hasn't been reported by its manufacturer. The other system is called 'CatWalk' \cite{PaperCatWalk}. It is comprised of a plexiglass walkway which can reflect light internally. When the animals paws touch the glass, the light escapes as their paw print and is captured by a high speed camera mounted  beneath the walkway.It can be used to quantize a number of gait parameters such as pressure, stride length, swing and stance duration. Multiple researchers have used 'CatWalk' in gait analysis \cite{PaperCatWalk1,PaperCatWalk2,PaperCatWalk3,PaperCatWalk4,PaperCatWalk5}.

\section{Hardware based methods}
\par \label{DrosophilaHardware} Kain et. al \cite{PaperKain} proposed an explicit hardware based leg tracking method for automated behavior classification in Drosophila flies.The fly is made to walk on a spherical treadmill. Dyes which are sensitive to specific wavelengths of light are applied to its legs and then the leg movement is recorded by two mounted cameras. This way, 15 gait features are recorded and tracked in real time. This approach has the appeal for real time deployment but it cannot be generalized to any limb tracking application because it needs a specific hardware setup. Moreover, being heavily dependent on photo-sensitive dyes decreases its robustness.

\par Snigdha et al \cite{PaperRoy} proposed 3D tracking of mice whiskers using optical motion capture hardware. The 3D tracking system used (Hawk Digital Real Time System, Motion Analysis Corp., Santa Rosa, CA, USA) is composed of two cameras in conjunctions and the Cortex analysis software (Motion Analysis, CA, USA). The whiskers are marker with retro-reflective markers and their X, Y, and Z coordinates are digitized and stored along with video recordings of the marker movements. The markers are fashioned from a retro-reflective tape backed with adhesive (Motion Analysis Corp., Santa Rosa, CA, USA) and fastened onto the whiskers using the tape’s adhesive. Markers were affixed to the whisker at a distance of about 1 cm from the base. Reliable 3D tracking requires that a marker be visible at all times by both cameras. This condition can be satisfied in head-fixed mice where the orientation of the mouse to the cameras remains fixed. The system was connected to a dual processor Windows based computer for data collection. The proposed tracking framework is easy to install and computationally cheap but like other hardware-assisted frameworks, it also needs specialized hardware and thus isn't very scalable and portable. Also, for reliable tracking, the retro-reflective markers should be visible to the cameras at all times which makes the framework less robust.

\par Scott Tashman et. al. proposed a bi-plane radiography assisted by static CT scan based method for 3D tracking of skeletons in small animals \cite{PaperScott}. The high-speed biplane radiography system consists of two 150 kVp X-ray generators optically coupled to synchronized high-speed video cameras. For static radiostereometric analysis [RefRSA] (RSA),they implanted minimum three radiopaque bone markers per bone to enable accurate registration between the two views. The acquire radiographs are first corrected for geometric distortion. They calculated ray-scale weighted centroids for each marker with sub-pixel resolution. They tested this system on dogs and reported an error of 0.02 mm when inter-marker distance calculated by their system was compared to true inter-marker distance of 30 mm. For dynamic gait tracking, this system is reported to be very accurate but it required specialized hardware. Moreover, since the marker implantation is invasive, it can alter the behavior of animals being studied.

\par Harvey et. al. proposed an optoelectronic based whisker tracking method for head-bound rats. In the proposed method, the rat’s head is fixed to a metal bar protruding from the top of the restraining device \cite{PaperHarvey}. Its paw rests on a micro switch which records lever presses. A turntable driven by a stepping motor rotates a single sphere/cube into the rat’s ¨whisking space¨. The whiskers are marked to increase chances of detection. The movements of a single whisker are detected by a laser emitter and an array of CCD detectors. Once the data is recorded, a single whisker is identified manually which serves as a reference point. As the article is more focused on whisking responses of the rodents to external stimuli, they have not reported the whiskers detection and tracking accuracy. R. Bermejo et. al reported similar approach for tracking individual whiskers \cite{PaperBemejo}. They restrained the rats and then used a combination of CCDs and laser emitters. The rats were placed in such a way that their whiskers blocked the path of laser, casting a shadow over CCDs, thus registering presence of a whisker which can be tracked by tracking the voltage shifts on CCD array. They also have not reported tracking accuracy. 

\par Kyme et. al. \cite{PaperKyme} proposed a marker assisted hardware based method for head motion tracking of freely behaving and tube-bounds rats. They glued a marker with a specific black and white pattern to the rat's head. Motion tracking was performed using the Micron-Tracker $S \times 60$ (ClaronTech. Inc., Toronto, Canada), a binocular-tracking system that computes a best-fit pose of printed markers in the field of measurement Kyme et al.\cite{PaperKymeMarker}. The author have reported accurate tracking for more than 95\% of the time in case of tube-bound rats and similar performance for freely behaving rats if the tracking algorithm is assisted 10\% of the time. These figures seems impressive but the approach has one major drawback; it can only be used in a very specific setting. It requires a specialized setup and it needs to glue external markers to the test subject's head, which might affect its behavior. Moreover, the same authors have used the Micron-Tracker based approach for synchronizing head movements of a rat with position emission tomography scans of their brains and have reported that the marker-assisted tracking method was able to synchronize the head movements with scan intervals with an error of less than 10 $ms$ \cite{PaperKymeSynch}.

\par Pasquet et. al. proposed a wireless inertial sensors based approach for tracking and quantifying head movements in rats \cite{PaperPasquet}. The inertial measurement unit (IMU) contains a digital 9-axis inertial sensor (MPU-9150, Invensense) that samples linear acceleration, angular velocity and magnetic field strength in three dimensions, a low-power programmable microcontroller (PIC16, Microchip) running a custom firmware and a Bluetooth radio, whose signal is transmitted through a tuned chip antenna. This system was configured with Labview for data acquisition and the analysis was done in R. The sensors record any head movements by registering the relative change in acceleration with respect to gravity. Since the sensors record data in 9 axes, it is used to detect events in rats behavior based on head movements. The authors have reported a detection accuracy of 96.3\% and a mean correlation coefficient of 0.78 $\pm$  0.14 when the recorded data is compared for different rats(n  =  19 rats). The reported performance figures are very good in terms of event detection and consistency but the system can only be used to track head movements. Also, the system requires specialized hardware which limits its portability.

\section{Video tracking aided by hardware}
\subsection{Semi-automated}
\par Knutsen et al. proposed the use of overhead IR LEDs along with normal video cameras for head and whisker tracking of unrestrained behaving mice \cite{PaperKnutsen}. The overhead IR leds are used to flash IR light onto the mouse head which is reflected back from its eyes. The reflected flash is recorded by an IR camera. In first few frame of every movie, a user identifies a region of interest (ROI) for the eyes which encircles a luminous spot (reflection from the eye). This luminous spot is tracked in subsequent frames by looking for pixels with high luminosity in the shifted ROI. Once eyes are located in every frame, they are used to track head and whiskers in intensity videos. First a mask averaged on all those frames which contains no mice is subtracted from the frame. Then user-initiated points are used to form whisker shaft by spline interpolation. For the next frame, sets of candidate points are initiated and shaft from current frames is convolved with candidate shafts from next frame to locate the set of points most likely being a whisker. Although the pipeline has no temporal context involved, yet it is quite efficient in whisker tracking with a high Pearson correlation between ground truth and tracked whisker shafts. The downsides of this approach are the need for high speed videos and additional IR hardware.

\par Gravel et. al. proposed an X-Ray area scan camera assisted tracking method for gait parameters of rats walking on a treadmill \cite{PaperGravel}. The system consists of a Coroskop C arm X-ray system from Siemens, equipped with an image intensifier OPTILUX 27HD. The X-Ray system is used in detecting flouroscopic markers placed on hind limbs of the rat. A a high-speed area scan camera from Dalsa (DS-41-300K0262), equipped with a C-mount zoom lens (FUJINON-TV, H6X12.R, 1:1.2/12.5–75) mounted on the image intensifier is used for video acquisition and a computer is used to overlay the detected markers on the video. The treadmill with the overlying box is placed on a free moving table and positioned near the X-ray image intensifier. The X-ray side view videos of locomotion are captured while the animal walked freely at different speeds imposed by the treadmill. The acquired video and marker data is processed in four steps; correction for image distortion, image denoising and contrast enhancement, frame-to-frame morphological marker identification and  statistical gait analysis. The data analysis process can be run in automated mode for image correction and enhancement however the morphological marker identification is user assisted.The kinematic gait patterns are computed using a Bootstrap method \cite{PaperBootstrap}. Using Bootstrap method and multiple Monte Carlo runs, the authors have reported consistent gait prediction and tracking with a confidence of 95\%. They have compared the performance of the proposed system with manual marker annotation by a user by first manually processing 1 hour 30 minutes of data and then processing only 12 minutes data by the system assisted by the same user. They have reported only 8\% deviation in gait cycle duration, therefore claiming a 7-folds decrease in processing time with acceptable loss in accuracy. Although the proposed results are impressive, the system is still not scalable and portable because it relies on dedicated hardware as well as continuous user assistance.

\subsection{Completely automated}
\par Akihiro Nakamura et. al. \cite{PaperAkihiro} proposed a depth sensor based approach for paw tracking of mice on a transparent floor. The proposed system captures the subjects shape from beneath using a low-cost infrared depth sensor (Microsoft Kinect) and an opaque infrared pass filter. The system is composed of an open-field apparatus, a Kinect sensor, and a personal computer. The open field is a square of $400 mm \times 400 mm$ and the height of the surrounding wall is 320 mm. The Kinect device is fixed 430 mm below the floor so that the entire open-field area can be captured by the device. For the experiment in the opaque conditions, the floor of the open field was covered with tiled infrared-pass filters (FUJIFILM IR-80 (Fuji Film, Tokyo, Japan)), which are commonly used in commercial cameras. The depth maps, consisting of $320 \times 240 $depth pixels, are captured at 30 frames per second. The tracking algorithm has four steps; preprocessing, feature-point extraction, footprint detection and labeling. During preprocessing, the subject’s depth information is extracted from the raw depth map by applying background subtraction to the raw depth map. The noise produced by pre-processing steps is removed by morphological operations. AGEX algorithm \citep{PaperAGEX} is used for feature extraction after pre-processing. Center of mass of AGEX point clouds is used for paw detection and labeling. All those pixels whose Euclidean distance is lower than a threshold from the center of mass are considered to be member pixels of the paws. This framework offers the benefits of low computational cost but it is not robust. It can be used only for paw tracking in a specific setting. 

\par César S. Mendes et. al. \cite{PaperMendes} proposed an integrated hardware and software system called 'MouseWalker' that provides a comprehensive and quantitative description of kinematic features in freely walking rodents. The MouseWalker apparatus primarily comprises four components: the fTIR floor and walkway wall, the supporting posts, the $45^0$ mirror, and the background light. A white LED light strip for black and white cameras or a colored LED light strip for color cameras is glued to a 3/8-inch
U-channel aluminum base LED mount. This LED/aluminum bar is clamped to the long edges of a 9.4-mm (3/8-inch) thick piece of acrylic glass measuring 8 by 80 cm. A strip of black cardboard is glued and sewn over the LED/acrylic glass contact areas. To build the acrylic glass walkway, all four sides were glued together with epoxy glue and cable ties and placed over the fTIR floor. Videos are acquired using a Gazelle 2.2-MP camera (Point Grey, Richmond, Canada) mounted on a tripod and connected to a Makro-Planar T 2/50 lens (Carl Zeiss, Jena, Germany) at maximum aperture (f/2.0) to increase light sensitivity and minimize depth of field. The 'MouseWalker' program is developed and compiled in MATLAB (The Mathworks, MA, USA) \cite{MouseWalker}. The body and footprints of the mouse are distinguished from the background and from each other based on their color or pixel intensity. The RGB color of the mouse body and footprints are user defined. The tail is identified as a consecutive part of the body below a thickness threshold. Three equidistant points along the tail are used to characterize tail curvature. Head is defined by the relative position of the nose. The center and direction of this head part are also recorded along with the center of the body without the tail and its orientation. A body "back" point is defined as the point which is halfway between the body center and start of the tail. For the footprints of the animal, the number of pixels within a footprint, as well as the sum of the brightness of these pixels, are stored by the software. The 'MouseWalker' can be used to track speed, steps frequency, swing period and length of steps, stance time, body linearity index footprint clustering and leg combination indexes: no swing, single-leg swing, diagonal-leg swing, lateral-leg swing, front or hind swing, three-leg swing, or all-legs swing (unitless). Like other hardware assisted methods, this method also suffers from the lack of portability and scalability.

\par Wang et al. proposed a pipeline for tracking motion and identifying micro-behavior of small animals based on Microsoft Kinect sensors and IR cameras \cite{PaperWang}. This is achieved by employing Microsoft Kinect cameras along with normal video cameras to record movement of freely behaving rodents from three different perspectives.The IR depth images from Microsoft Kinect are used to extract shape of the rodents by background subtraction. After shape extraction, five pixel-based features are extracted from the resultant blobs which are used for tracking and behavior classification by Support Vector Machines. Although the pipeline is not exclusively used for motion tracking, yet the idea of using depth cameras is potentially a good candidate for motion tracking as well.

\par Monteiro et al. \cite{PaperMoteiro} took a similar approach to Wang et al. \cite{PaperWang} by using Microsoft Kinect depth cameras for video capturing \cite{PaperMoteiro}. Instead of using background subtraction, they introduced a rough temporal context by tracking morphological features of multiple frames for motion tracking. In their approach, the morphological features are extracted frame by frame. Then features from multiple adjacent frames are concatenated to introduce a rough temporal context. Finally decision trees are used for behavior classification. A decision tree is then trained from this dataset for automatic behavior classification. The authors have reported a classification accuracy of 66.9\% when the classifier is trained to classify four behaviors on depth map videos of 25 minutes duration. When only three behaviors are considered, the accuracy jumps to 76.3\%. Although the introduced temporal context is rough and the features are primitive, the classification performance achieved firmly establishes the usefulness of machine learning in behavioral classification. Like \cite{PaperMoteiro}, this approach is also not solely used for motion tracking, but they have introduced a rough temporal context for tracking along with depth cameras which can be beneficial in motion tracking only approaches.

\par Voigts et al. proposed an unsupervised whisker tracking pipeline aided by the use of IR sensors for selective video capturing \cite{PaperVoigts}. They captured high speed (1000 frames per seconds) video data by selectively recording those frames which contained mice. It was achieved by sensing the mice by an IR sensor which then triggers the video camera to start recoding. Once the mice leaves the arena, the IR sensors triggers the video camera to stop capturing. This selectively-acquired video data is used for whisker tracking. First, a background mask is calculated by averaging 100 frames containing no mice. This mask is subtracted from every single frame. Then vector fields from each frame that resulted in a convergence of flows on whisker-like structures are generated. These fields are then integrated to generate spatially continuous traces of whiskers which are grouped into whisker splines. This approach is completely unsupervised when it comes to whisker tracking with a rough temporal context as well but it is very greedy in terms of computational resources so it cannot be employed in real time.

\par Petrou et. al. \cite{PaperPetrou} proposed a marker-assisted pipeline for tracking legs of female crickets. The crickets are filmed with three cameras, two mounted above and one mounted below the crickets which are made to walk on transparent glass floor. Leg joints are marked with fluorescent dyes for better visualization. The tracking procedure is initiated by a user by selecting marker position in initial frames. The initial tracking is carried out to next frames by constrained optimization and using euclidean distance between joints of current frame and the next frame. This pipeline does a decent job in terms of tracking performance as the average deviation between human annotated ground truth (500 digitized frames) and automatic tracking is 0.5 mm where the spatial depth of the camera is 6 pixel/mm. This approach however requires special setup and cannot be exported to other environments.

\par Xu et. al. \cite{PaperXu} proposed another marker assisted tracking pipeline for small animals. In proposed pipeline, the limbs and joints are first shaved, marked with dyes and then recorded with consumer grade cameras (200 frames per second). Tracking is then done in steps which include marker position estimation, position prediction and mismatch occlusion. Marker position is estimated by correlation in two methods. In one method, normalized cross correlation between gray scale region of interest and user generated sample markers is found. The pixels with highest correlation are considered as the marker pixels. In the second method, normalized covariance matrix of marker model and color ROI is used to estimate pixels with highest normalized covariance values which are considered as marker pixels. Once the marker positions are estimated in current frame, they are projected to next frame by polynomial fitting and Kalman filers. For occlusion handling, they assume that a marker position or image background cannot change abruptly, so if there is a sudden change, it must be an occlusion. The approach is simple and scalable enough to be exported to any environment while at the same time, due to its dependency on markers, its not robust. 

\par John et. al. \cite{PaperJohn} proposed a semi-automated approach for simultaneously extracting three-dimensional kinematics of multiple points on each of an insect’s six legs. White dots are first painted on insects leg joints. Two synchronized video cameras placed under the glass floor of the platform are used to record video data at 500 frames per second. The synchronized video data is then used to generate 3D point clouds for the regions of interest by triangulation. The captured video frames are first subtracted from a background frame modeled by a Gaussian mean of 100 frames with no insects. After image enhancement, a user defines the initial tracking positions of leg joints in 3D point cloud which are then tracked both in forward and backward direction automatically. The user can correct any mismatched prediction in any frame. The authors have reported a tracking accuracy of 90\% when the user was allowed to make corrections in 3-5\% of the frames. Proposed approach is simple in terms of implementation, accurate in terms of spatial and temporal resolution and easy to operate. However, it needs constant user assistance and does not have any self-correction capability.

\par Hwang et. al. \cite{PaperHwang} followed a similar approach to the one proposed by John et. al. \cite{PaperJohn} but without the use of markers. They used a combination of six-color charge-coupled device (CCD) cameras (BASLER Co. Sca640-70fc) for video recording of the insects. To capture the diverse motions of the target animal, they used two downward cameras and four lateral cameras as well as a transparent acrylic box. The initial skeleton of the insect was calculated manually, so the method is not completely automated. After the initial skeleton, they estimated the roots and extremities of the legs followed by middle joints estimation. Any errors in the estimation were corrected by Forward And Backward Reaching Inverse Kinematics (FABRIK) \cite{PaperFABRIK}. The authors have not reported any quantitaive results which might help us to compare it with other similar approaches however they have included graphics of their estimation results in the paper. This paper does not directly deal with motion estimation in rodents, however, given the unique approach to using cameras and pose estimation, it is a worthwhile addition to the research in the field.

\section{Video tracking methods mostly dependent on software based tracking}
In this section, we will focus on all those research works which try to solve the locomotion and gesture tracking problem by processing raw and un-aided video streams. In this scenario, there is no specialized hardware installed apart from one or multiple standard video camera. There are no physical markers on the mice/animals bodies as well which can help track its motion. These works approaches the problem from purely a computer vision point of view.
\subsection{Semi-automated}
\par Gyory et. al \cite{PaperGyory} proposed a semi-automated pipeline for tracking rat's whiskers. In proposed pipeline, videos are acquired with high speed cameras (500 frames per second) and are first pre-processed to adjust its brightness.The brightness adjusted image is eroded to get rid of small camera artifacts. Then a static background subtraction is applied which leaves only the rat body in the field of view. As whiskers are represented by arcs with varying curvature, a polar-rectangular transform is applied and then a horizontal circular shift is introduced so that whiskers are aligned as straight lines on a horizontal plane. Once the curved whiskers are represented by straight lines, hough transform is used to locate them.The approach is too weak and non robust to be considered for any future improvements. The reported computational cost is high (processing speed of 2 fps). Also, it works on high speed videos (>500 fps). It is highly sensitive to artifacts and it cannot take care of occlusion, dynamic noise and broken whisker representation.

\par Hamers et. al. proposed a specific setup based on inner-reflecting plexi-glass walkway \cite{PaperHamers}.The animals traverse a walkway (plexi-glass walls,
spaced 8 cm apart) with a glass floor (109 3 15 3 0.6
cm) located in a darkened room. The walkway is illuminated by a fluorescent tube from long edge of the glass floor. For most of the way, the light travels internally in the glass walkway, but when some pressure is applied, for example by motion of a mouse, the light escapes and is visible from outside. The escaped light, which is scattered from the paws of the mouse, is recorded by a video camera aimed at a $45^0$ mirror beneath the glass walkway. The video frames are then thresholded to detect bright paw prints. The paws are labeled (left, right, front, hind). The system can extrapolate a tag (label of the footprint) to the bright areas in next frame which minimizes the need for user intervention but in some cases, user intervention becomes necessary. The authors haven't reported paw detection/tracking performance.

\subsection{Completely automated}
\par Da Silva et. al. conducted a study on the reproducibility of automated tracking of behaving rodents in controlled environments \cite{PaperDaSilva}. rats in a circular box of 1 m diameter with 30 cm walls. The monitoring camera was mounted in such a way that it captured the rodents from top view while they were behaving. They used a simple thresholding algorithm to determine pixels belonging to the rodent. Athough the method is rudimentary as compared to state of the art, the authors have reported a pearson correlation of $r = 0.873$ when they repeated the same experiment at different ages of the animals, thus validating its reproducibility. However, this setup can only be used to track whole body of rodents, it cannot identify micro-movements such as limbs motion.

\par Leroy et. al. proposed the combination of transparent Plexiglas floor and background modeling based motion tracking \cite{PaperLeroy}. The rodents were made to walk on a transparent plexiglas floor illuminated by florescent light and was recorded from below. A background image was taken when there was no mouse on the floor. This background image was subtracted from every video frame to produce a continuously updating mouse silhouette. Mouse tail was excluded by an erosion followed by dilation of the mouse silhouette. Then the center of mass of the mouse was calculated which was tracked through time to determine if the mouse is running or walking. Since the paws are colored, color segmentation is used to isolate paws from the body. The authors have reported a maximum tracking error of $4 \pm 1.9$ and a minimum tracking error of $2 \pm 1.6$ when 203 manually annotated footprints are compared to their automatic counter parts.   

\par Dankert et. al. proposed a machine vision based automated behavioral classification approach for Drosophila \cite{PaperDankert}. The approach does not cover locomotion in rodents, it covers micro-movements in flies. Videos of a pair of male and female flies are recorded for 30 minutes in a controlled environment. Wing beat and legs motion data is manually annotated for lunging, chasing, courtship and aggression. The data analysis consists of four stages. In first stage, Foreground image $F_I$ computed by dividing the original image I by ($\mu I + 3\sigma_I$) ($F_I$ values in false-colors). In second stage, The fly body is localized by fitting a Gaussian mixture model \cite{PaperGMM} (GMM) with three Gaussians; background, other parts and body to the histogram of $F_I$ values (gray curve) using the Expectation Maximization (EM) algorithm \cite{PaperGMM}. First (top) and final (bottom) iterations of the GMM-EM optimization. All pixels with brightness values greater than a threshold are assigned to the body, and are fit with an ellipse. In third stage, full fly is detected by segmenting the complete fly from the background, with body parts and wings \cite{PaperOtsu}. In fourth stage, head and abdomen are resolved by dividing the fly along the minor axis of the body ellipsoid and comparing the brightness-value distribution of both halves. In fifth stage, 25 measurements are computed, characterizing body size, wing pose, and position and velocity of the fly pair. A k-nearest neighbor classifier is trained for action detection. The authors have reported a false positive rate for lunging at 0.01 when 20 minutes worth of data was used for training the classifier. Although this article does not directly deal with rodents, the detection and tracking algorithms used for legs and wings can be used for legs motion detection in rodents too.

\par Nathan et al. \cite{PaperNathan} proposed a whisker tracking method for mice based on background subtraction, whisker modeling and statistical approaches. Head of the mice as fixed, so they were not behaving freely. They used a high speed camera with a shutter speed of 500 frames per second. In order to track whiskers, an average background image was modeled from all the video frames and then subtracted from every single frame. Afterwards, pixel level segmentation was done to initiate candidate sites by looking for line like artifacts. Once the candidate boxes are initiated, they are modeled by two ellipsoids with perpendicular axes. The ellipsoid with higher eccentricity is the best possible candidate site for whiskers. These whiskers are then traced in every single frame of the video sequence by using expectation maximization. The approach has some strong points. It requires no manual initiation, it is highly accurate and because of superb spatial resolution and pixel-level tracking, even micro-movements of whiskers can be tracked. But all the strengths come at a cost; the approach is computationally very expensive which means it cannot be deployed in real time. There is another downside to pixel-level and frame-level processing, the temporal context is lost in the process.

\par Kim et al. \cite{PaperKim} proposed a method similar to the one proposed by Clack. et al. \citep{PaperNathan} to track whisker movements in freely behaving mice. They use Otsu's algorithm to separate foreground and background and then find the head of the mouse by locating triangular shaped object in the foreground. Once the head and snout are detected, hough transform is used to find line-like shapes (whiskers) on each side of the snout. Mid points of the detected lines are used to form ellipsoidal regions which help track whiskers in every single frame. This pipeline was proposed to track whisking in mice after a surgical procedures. There is no ground truth available, so the approach cannot be evaluated for tracking quantitatively. Besides, the pipeline is not feasible for real time deployment due to high computational costs.

\par Palmer et al. proposed a paw-tracking algorithm for mice when they grab food and can be used for gesture tracking as well \citep{PaperPalmer9}. They developed the algorithm by treating it as a pose estimation problem. They model each digit as a combination of three phalanges (bones). Each bone is modeled by an ellipsoid. For 4 digits, there are total 12 ellipsoids. The palm is modeled by an additional ellipse. Forearm is also modeled as an ellipsoid while nose is modeled as an elliptic paraboloid. The  paw is modeled using 16 parameters for the digits (four degrees of freedom per digit), four constant vectors representing the metacarpal bones and 6 parameters for position and rotation of the palm of the paw. Furthermore, the forearm is assumed to be fixated at the wrist and can rotate along all three axes in space. This amounts to a total of 22 parameters. In each frame, these ellipsoids are projected in such away that they best represent the edges. The best projection of ellipsoids is found by optimization and is considered a paw. They haven't reported any quantitative results. This approach is very useful if the gesture tracking problem is treated as pose estimation with a temporal context.

\par In \cite{PaperPalmer}, Palmer et al. extended their work from \cite{PaperPalmer9}.The basic idea is the same. It models the paw made of different parts. Four digits (fingers), each digit having 3 phalanges (bones). Each phalanges is modeled by an ellipsoid, so total 12 ellipsoid for the phalanges plus an additional one for the palm. In this paper, the movement of the 13 ellipsoids is modeled by 19 degree freedom vectors, unlike 22 from \cite{PaperPalmer9}. The solution hypothesis is searched not simultaneously, but in stages to reduce the number of calculations. This is done by creating different number of hypotheses for every joint of every digit and then finding the optimum hypotheses. $\backslash$

\par A Giovannucci et. al. \cite{PaperOurs} proposed an optical flow and cascade learners based approach for tracking of head and limb movements in head-fixed mice walking/running on spherical/cylindrical treadmill. Unlike other approaches, only one camera installed from a lateral field of view was used for limbs tracking and one camera installed in front of the mouse was used for whisker tracking. The calculated dense optical flow fields in frame-to-frame method for whisker tracking. The estimated optical flow fields were used to train dictionary learning algorithms for motion detection in whiskers. They annotated 4217 frames for limbs detection and 1053 frames for tails detection and then used them to train Haar-Cascades classifiers for both the cases. They have reported a high correlation of $0.78 \pm 0.15$ for whiskers and $0.85 \pm 0.01$ for hind limb. The proposed hardware solution in the paper is low cost and easy to implement. The tracking approach is also computationally not demanding and can be run in real time. They however did not deal with the micro-patterns in motion dynamics which can be best captured with the inclusion of temporal context to the tracking approach.  

\par Heidi et. al, proposed  Automated Gait Analysis Through Hues and Areas (AGATHA) \cite{PaperAGATHA}. AGATHA first isolates the sagittal view of the animal by subtracting a background image where the animal is not present, transforming the frame into a HSV (Hues, Saturation, Value) image. The hue values are used to convert the HSV image into a binomial silhouette. Next, AGATHA locates the row of pixels representing the interface between the rat and the floor. AGATHA may not accurately locate the rat-floor interface if the animal moves with a gait pattern containing a completely aerial phase. Second, AGATHA excludes the majority of nose and tail contacts with the floor by comparing the contact point to the animal’s center of area in the sagittal view. Foot contact with the ground is visualized over time by stacking the rat/floor interface across multiple frames. The paw contact stacked over multiple frames is then used for gait analysis. Multiple gait parameters such as limbs velocity, stride frequency can be calculated. When results from AGATHA were compared to manual annotation on a 1000 fps video, they deviated by a small amount. For example, limbs velocity calculated b AGATHA was 1.5\% off from the velocity calculated manually. Similarly, AGATHA registered a difference of 0.2 cm in stride length from manual annotation. In general, the approach is simple and scalable but limited in scope.

\section{Conclusion}
\par The gesture detection and tracking approaches are still in the developing phase. There is no single approach powerful enough which can track micro-movements of limbs, whiskers or snout of the rodents which are necessary for gesture identification and behavioral phenotyping. In general, those approaches which use specialized hardware are more successful than those approaches which solely depend on standard video camera. For example, the use of X-Ray imaging to detect surgically implanted markers has been proven very successful to track limbs and joins movements with high precision. Moreover, the use of specific markers attached to either limbs or whiskers of the rodents also increase the overall tracking accuracy of an approach. However, there is a downside to this approach, the rodents might not behave naturally. Therefore, more and more research is being conducted on scalable, portable and non invasive tracking methods which only need standard video cameras. 
\par We have summarized some important aspects of selected approaches in table \ref{table1}. Following things need to be kept in mind To properly interpret the table.
\paragraph{Code availability:} It means whether the code is available or not. If it is available, is it free or paid.
\paragraph{Performance:} If the performance is given in terms of standard deviation, it signifies the consistency of proposed approach either against itself or an annotated dataset (which is pointed out). For example, if the table says that the proposed system can make a 90\% accurate estimation of limbs velocity with an SD of 3\%, it means that the system performance fluctuates somewhere between 87\% to 93\%. If absolute accuracy is given, it means each and every detected instant is compared to manually annotated samples. If only \% SD is given or just SD is given, it means that the system can consistently reproduce the same result with specified amount of standard deviation, regardless of its performance against the ground truth.
\paragraph{Need specialized setup \& Invasiveness}: This means that whether the method need any specialized hardware other than the housing setup or video cameras. If they housing setup itself is arranged in a specific way but it does not contain any specialized materials, we say that the hardware setup required is not specialized. By invasiveness, we mean that a surgery has to be conducted to implant the markers. If no surgery is needed to implant markers, we call it semi-invasive. If no markers are needed, we call it non-invasive.

\begin{landscape}
\begin{longtable} 
{|p{0.5cm}|l|p{4cm}|p{9cm}|p{2cm}|p{2cm}|}

\caption{Comparison of different approaches. Legend:: Invasive: Approaches which requires surgery to put markers for tracking, semi-invasive: Approaches which do not need surgery for marker insertion, non-invasive: no marker needed. Real time means that the system can process frames at the same rate they are being acquired. If it needs specialized equipment apart from standard video cameras and housing setup, it is pointed out in the last column}\\
\label{table1}\\
\hline
      & Type  & Code availability & Performance & Real time or offline & Need specialized setup \& Invasiveness \\
    \hline
    \cite{PaperDigiGaitAndTreadScan}& Commercial & Paid  & Comparison with ground truth not provided. One paper reports the reproducibilty: 2.65 \% max SD  & Yes   & Yes \\
     \hline
    \cite{PaperGaitScan} & Commercial & Paid  & Comparison with ground truth not provided. One paper reports the reproducibilty: 1.57 \% max SD  & Yes   & Yes \\
    \hline 
    \cite{PaperKain} & Research & data and code for demo available at http://lab.debivort.org/leg-tracking-and-automated-behavioral-classification-in-Drosophila/ & tracking performance not reported, behavioral classification of 12 traits reported to be max at 71\% & Tracking real time, classification offline & yes \\
    \hline 
    \cite{PaperScott} & Research & not available & tracking: SD of only 0.034\% when compared with ground truth, Max SD of 1.71 degrees in estimating joint angle & real time legs and joints tracking & yes, invasive \\
    \hline 
    
    \cite{PaperHarvey} & Research & not available & tracking performance not reported explicitly & real time whisker tracking & yes, semi-invasive \\
    \hline 
    \cite{PaperBemejo} & Research & available on request & whisker tracking performance not reported explicitly & real time single whisker tracking & yes, semi-invasive \\
    \hline 
    \cite{PaperKyme} & Research & not available & head motion tracked correctly with a max false postive of 13\% & real time head and snout tracking & yes, semi-invasive \\
    \hline 
    \cite{PaperKymeMarker} & Research & not available & head motion tracked continuously with a reported SD of only 0.5 mm & real time head and snout tracking & yes, semi-invasive \\
    \hline 
    \cite{PaperPasquet} & Research & not available & head motion tracked with an accuracy of 96.3 \% and the tracking can be reproduced over multiple studies with a correlation cefficient of 0.78 & real time head tracking & yes, semi-invasive \\
    \hline 
    \cite{PaperKnutsen} & Research & code and demo data available at https://goo.gl/vYaYPy & they reported a correlation between whisking amplitude and velocity as a measure of reliability, R = 0.89 & Offline head and whisker tracking & no, invasive \\
    \hline 
    \cite{PaperGravel} & Research & not available & Tracking and gait prediction with confidence of 95 \%, deviation between human annotator and computer at 8\%  & Offline & yes, semi-invasive \\
    \hline
    
        \cite{PaperAkihiro} & Research & not available & Paw tracked with an accuracy of 88.5 on transparent floor and 83.2 \% on opaque floor  & Offline & yes, semi-invasive \\
    \hline
    
     \cite{PaperMendes} & Research & code available at \url{https://goo.gl/58DQij} & tail and paws tracked with an accuracy >90 \%  & Real time & yes, semi-invasive \\
    \hline
    
         \cite{PaperWang} & Research & not available & 5 class behavioral classification problem, accuracy in bright condition is 95.34 and in dark conditions is 89.4\%  & offline & yes, non-invasive \\
    \hline
    
    \cite{PaperMoteiro} & Research & not available & 6 behavioral class accuracy: 66.9 \%, 4 behavioral class accuracy: 76.3\%  & offline & yes, non-invasive \\
    \hline
    
        \cite{PaperVoigts} & Research & code available at \url{https://goo.gl/eY2Yza} & whisker detection rate: 76.9\%, peak spatial error in whisker detection: 10 pixels & offline & yes, non-invasive \\
    \hline
    
       \cite{PaperPetrou} & Research & not available & Peak deviation between human annotator and automated annotation: 0.5 mm with a camera of 6 pixel/mm resolution & offline & yes, non-invasive \\
    \hline
    
           \cite{PaperJohn} & Research & not available & Tracking accuracy >90 \% after the algorithm was assisted by human users in 3-5 \% of the frames & offline & yes, semi-invasive \\
    \hline
    
       \cite{PaperGyory} & Research & code available at \url{https://goo.gl/Gny89o} & A max deviation of 17.7\% between human and automated whisker annotation & offline & yes, non-invasive \\
    \hline
    
       \cite{PaperLeroy} & Research & not available & Maximum paw detection error: 5.9 \%, minimum  error : 0.4 \%  & offline & no, non-invasive \\
    \hline
    
           \cite{PaperDankert} & Research & Source code at \url{https://goo.gl/zesyez}  , demo data at \url{https://goo.gl/dn2L3y} & Behavioral classification: 1\% false positive rate & offline & no, semi-invasive \\
    \hline
    
           \cite{PaperNathan} & Research & Source code available at \url{https://goo.gl/JCv3AV} & Whisker tracing accuracy: max error of 0.45 pixels & offline & no, non-invasive \\
    \hline
    
           \cite{PaperOurs} & Research & not available & Correlation with annotated data; for whiskers r = 0.78, for limbs r = 0.85 & real time& no, non-invasive \\
    \hline
    
            \cite{PaperAGATHA} & Research & code available at \url{https://goo.gl/V54mpL} & Velocity calculated by AGATHA was off from manually calculated velocity by 1.5\% & real time& no, non-invasive \\
    \hline
    
    \end{longtable}
\bigskip\centering
\end{landscape}

\subsection{Future Research}
\par Based on the literature survey we conducted, we have the following recommendations for future research:
\begin{enumerate}
\item Methods would benefit from an effective use of different camera configurations to get spatial data at high resolution in 3D space.
\item One of the most relevant shortcomings of the field is the lack of public databases to validate new algorithms. Different approaches are tested on the (usually private) data from the lab developing the solution. Building a standardized gesture tracking dataset which can be used as a benchmark would benefit the community in a similar way as large object recognition databases (PASCAL, ImageNet or MS COCO) allowed significant progress in the Computer Vision literature.
\item Currently there exist large amounts of non labeled data samples (thousands of video hours). The use of unsupervised learning algorithms that could benefit the parameter learning of supervised methods is one of the most challenging future research lines.
\item In addition, the use of semi-supervised and weakly-supervised learning algorithms could benefit the community. The challenge in this particular case is to minimize the user intervention (supervision) maximizing the improvements on the accuracy.
\item Finally, deep learning methods have been shown to outperform many computer vision tasks. Their application to this field seems a promising research line. 

\end{enumerate}


\section{Bibliography}

\begin{thebibliography}{1}
\bibitem{BkgPaper1}
De-gui, Xiao, Yu Sheng-sheng, and Zhou Jing-li. "Motion tracking with fast adaptive background subtraction." Wuhan University Journal of Natural Sciences A 8.1 (2003): 35-40.

\bibitem{BkgPaper2}
Zhang, Ruolin, and Jian Ding. "Object tracking and detecting based on adaptive background subtraction." Procedia Engineering 29 (2012): 1351-1355.

\bibitem{BkgPaper3}
Saravanakumar, S., A. Vadivel, and CG Saneem Ahmed. "Multiple human object tracking using background subtraction and shadow removal techniques." Signal and Image Processing (ICSIP), 2010 International Conference on. IEEE, 2010.

\bibitem{BkgPaper4}
Kim, Intaek, Tayyab Wahab Awan, and Youngsung Soh. "Background subtraction-based multiple object tracking using particle filter." Systems, Signals and Image Processing (IWSSIP), 2014 International Conference on. IEEE, 2014.
APA	

\bibitem{BkgPaper5}
Zhang, Lijing, and Yingli Liang. "Motion human detection based on background subtraction." Education Technology and Computer Science (ETCS), 2010 Second International Workshop on. Vol. 1. IEEE, 2010.

\bibitem{BkgPaper6}
Shuigen, Wei, Chen Zhen, and Dong Hua. "Motion detection based on temporal difference method and optical flow field." Electronic Commerce and Security, 2009. ISECS'09. Second International Symposium on. Vol. 2. IEEE, 2009.

\bibitem{BkgPaper7}
Singla, Nishu. "Motion detection based on frame difference method." International Journal of Information \& Computation Technology 4.15 (2014): 1559-1565.

\bibitem{BkgPaper8}
Lu, Nan, et al. "An Improved Motion Detection Method for Real-Time Surveillance." IAENG International Journal of Computer Science 35.1 (2008).

\bibitem{BkgPaper9}
Jing, Guo, Chng Eng Siong, and Deepu Rajan. "Foreground motion detection by difference-based spatial temporal entropy image." TENCON 2004. 2004 IEEE Region 10 Conference. IEEE, 2004.

\bibitem{BkgPaper10}
Shaikh, Soharab Hossain, Khalid Saeed, and Nabendu Chaki. "Moving object detection approaches, challenges and object tracking." Moving Object Detection Using Background Subtraction. Springer, Cham, 2014. 5-14.

\bibitem{StatPaper1}
Denzler, Joachim, et al. "Statistical approach to classification of flow patterns for motion detection." Image Processing, 1996. Proceedings., International Conference on. Vol. 1. IEEE, 1996.

\bibitem{StatPaper2}
Denzler, Joachim, et al. "Statistical approach to classification of flow patterns for motion detection." Image Processing, 1996. Proceedings., International Conference on. Vol. 1. IEEE, 1996.

\bibitem{StatPaper3}
Paragios, Nikos, and George Tziritas. "Adaptive detection and localization of moving objects in image sequences." Signal Processing: Image Communication 14.4 (1999): 277-296.

\bibitem{StatPaper4}
Hu, Weiming, et al. "A system for learning statistical motion patterns." IEEE transactions on pattern analysis and machine intelligence 28.9 (2006): 1450-1464.

\bibitem{StatPaper5}
El Abed, Abir, Séverine Dubuisson, and Dominique Béréziat. "Comparison of statistical and shape-based approaches for non-rigid motion tracking with missing data using a particle filter." International Conference on Advanced Concepts for Intelligent Vision Systems. Springer, Berlin, Heidelberg, 2006.

\bibitem{StatPaper6}
Chellappa, Rama, et al. "Statistical methods and models for video-based tracking, modeling, and recognition." Foundations and Trends in Signal Processing 3.1–2 (2010): 1-151.

\bibitem{StatPaper7}
Paragios, Nikos, and Rachid Deriche. "Geodesic active contours and level sets for the detection and tracking of moving objects." IEEE Transactions on pattern analysis and machine intelligence 22.3 (2000): 266-280.

\bibitem{StatPaper8}
Pless, Robert, Tomas Brodsky, and Yiannis Aloimonos. "Detecting independent motion: The statistics of temporal continuity." IEEE transactions on pattern analysis and machine intelligence 22.8 (2000): 768-773.

\bibitem{StatPaper9}
Isard, Michael. Visual motion analysis by probabilistic propagation of conditional density. Diss. University of Oxford. 1998., 1998.

\bibitem{StatPaper10}
Comaniciu, Dorin, and Peter Meer. "Mean shift: A robust approach toward feature space analysis." IEEE Transactions on pattern analysis and machine intelligence 24.5 (2002): 603-619.

\bibitem{FlowPaper1}
Lucas, Bruce D., and Takeo Kanade. "An iterative image registration technique with an application to stereo vision." (1981): 674-679.

\bibitem{FlowPaper2}
Horn, Berthold KP, and Brian G. Schunck. "Determining optical flow." Artificial intelligence 17.1-3 (1981): 185-203.

\bibitem{FlowPaper3}
DWixson, L. "Detecting salient motion by accumulating directionally-consistent flow." IEEE transactions on pattern analysis and machine intelligence 22.8 (2000): 774-780.

\bibitem{FlowPaper4}
Shafie, Amir Akramin, Fadhlan Hafiz, and M. H. Ali. "Motion detection techniques using optical flow." World Academy of Science, Engineering and Technology 56 (2009): 559-561.

\bibitem{FlowPaper5}
Aslani, Sepehr, and Homayoun Mahdavi-Nasab. "Optical flow based moving object detection and tracking for traffic surveillance." International Journal of Electrical, Computer, Energetic, Electronic and Communication Engineering 7.9 (2013): 1252-1256.

\bibitem{FlowPaper6}
Barron, John L., David J. Fleet, and Steven S. Beauchemin. "Performance of optical flow techniques." International journal of computer vision 12.1 (1994): 43-77.

\bibitem{FlowPaper7}
Sun, Shuyang, et al. "Optical Flow Guided Feature: A Fast and Robust Motion Representation for Video Action Recognition." arXiv preprint arXiv:1711.11152 (2017).


\bibitem{PaperDigiGaitAndTreadScan}
Dorman, Christopher W., et al. "A comparison of $DigiGait^{TM}$ and $TreadScan^{TM}$ imaging systems: assessment of pain using gait analysis in murine monoarthritis." Journal of pain research 7 (2014): 25.


\bibitem{PaperDigiGait1}
Xiao, Jianfeng, et al. "Motor phenotypes and molecular networks associated with germline deficiency of Ciz1." Experimental neurology 283 (2016): 110-120.

\bibitem{PaperDigiGait2}
Connell, James W., Rachel Allison, and Evan Reid. "Quantitative gait analysis using a motorized treadmill system sensitively detects motor abnormalities in mice expressing ATPase defective spastin." PloS one 11.3 (2016): e0152413.

\bibitem{PaperDigiGait3}
Sashindranath, M., M. Daglas, and R. L. Medcalf. "Evaluation of gait impairment in mice subjected to craniotomy and traumatic brain injury." Behavioural brain research 286 (2015): 33-38.

\bibitem{PaperDigiGait4}
Neckel, Nathan D. "Methods to quantify the velocity dependence of common gait measurements from automated rodent gait analysis devices." Journal of neuroscience methods 253 (2015): 244-253.

\bibitem{PaperDigiGait5}
Lambert, C. S., et al. "Gait analysis and the cumulative gait index (CGI): Translational tools to assess impairments exhibited by rats with olivocerebellar ataxia." Behavioural brain research 274 (2014): 334-343.

\bibitem{PaperDigiGait6}
Takano, Morito, et al. "In vivo tracing of neural tracts in tiptoe-walking yoshimura mice by diffusion tensor tractography." Neuroprotection and Regeneration of the Spinal Cord. Springer, Tokyo, 2014. 107-117.

\bibitem{PaperDigiGait7}
Hampton, Thomas G., and Ivo Amende. "Treadmill gait analysis characterizes gait alterations in Parkinson's disease and amyotrophic lateral sclerosis mouse models." Journal of motor behavior 42.1 (2009): 1-4.

\bibitem{PaperTreadScan1}
Beare, Jason E., et al. "Gait analysis in normal and spinal contused mice using the TreadScan system." Journal of neurotrauma 26.11 (2009): 2045-2056.

\bibitem{PaperTreadScan2}
Gellhaar, S., et al. "Chronic L‐DOPA induces hyperactivity, normalization of gait and dyskinetic behavior in MitoPark mice." Genes, Brain and Behavior 14.3 (2015): 260-270.

\bibitem{PaperTreadScan3}
Beare, Jason. Kinematic analysis of treadmill walking in normal and contused mice using the TreadScan system. University of Louisville, 2007.

\bibitem{PaperTreadScan5}
McMackin, Marissa Z., Chelsea K. Henderson, and Gino A. Cortopassi. "Neurobehavioral deficits in the KIKO mouse model of Friedreich’s ataxia." Behavioural brain research 316 (2017): 183-188.

\bibitem{PaperTreadScan4}

\bibitem{PaperGaitScan}
\url{http://cleversysinc.com/CleverSysInc/csi_products/gaitscan/}

\bibitem{PaperGaitScan1}
E.B. Adamah-Biassi, I. Stepien, R.L. Hudson, M.L. Dubocovich (2013), Automated Video Analysis System Reveals Distinct Diurnal Behaviors in C57BL/6 and C3H/HeN Mice, Behavioural Brain Research, Volume 243, 15 April 2013, Pages 306–312.

\bibitem{PaperGaitScan2}
Evan J. Kyzar, Mimi Pham, Andrew Roth, Jonathan Cachat, Jeremy Green, Siddharth Gaikwad, Allan V. Kalueff, (2012) Alterations in grooming activity and syntax in heterozygous SERT and BDNF knockout mice: the utility of behavior-recognition tools to characterize mutant mouse phenotypes, Brain Research Bulletin, Dec 2012; 89(5-6): 168-76

\bibitem{PaperGaitScan3}
Ekue Bright Adamah-Biassi, Iwona Stepien, Randall L Hudson, and Margarita L Dubocovich, (2012) Effects of the Melatonin Receptor Antagonist (MT2)/Inverse Agonist (MT1) Luzindole on Re-entrainment of Wheel Running Activity and Spontaneous Homecage Behaviors in C3H/HeN Mice, FASEB J, Apr 2012; 26: 1042.5.

\bibitem{PaperGaitScan4}
Evan Kyzara, Siddharth Gaikwada, Andrew Rotha, Jeremy Greena, Mimi Phama, Adam Stewart, Yiqing Liang, Vikrant Kobla, Allan V. Kalueff (2011) a Towards high-throughput phenotyping of complex patterned behaviors in rodents: Focus on mouse self-grooming and its sequencing, Behavioural Brain Research (2011) Behavioural Brain Research Volume 225, Issue 2, 1 December 2011, Pages 426-431.

\bibitem{PaperGaitScan5}
Tai-Hsien Ou-Yang, Meng-Li Tsai, Chen-Tung Yen, Ta-Te Lin, An infrared range camera-based approach for three-dimensional locomotion tracking and pose reconstruction in a rodent, Journal of Neuroscience Methods Volume 201, Issue 1, 30 September 2011, Pages 116-123.

\bibitem{PaperELadder}
Cupido A: Detecting Cerebellar Phenotypes with the Erasmus Ladder.
Erasmus MC. Dept of Clin Genet Ph.D. 2009; 98.

\bibitem{PaperELadder1}
Ha, Seungmin, et al. "Cerebellar Shank2 regulates excitatory synapse density, motor coordination, and specific repetitive and anxiety-like behaviors." Journal of Neuroscience 36.48 (2016): 12129-12143.

\bibitem{PaperELadder2}
Peter, Saša, et al. "Dysfunctional cerebellar Purkinje cells contribute to autism-like behaviour in Shank2-deficient mice." Nature communications 7 (2016): 12627.

\bibitem{PaperELadder3}
De Zeeuw, Chris I., and Tycho M. Hoogland. "Reappraisal of Bergmann glial cells as modulators of cerebellar circuit function." Frontiers in cellular neuroscience 9 (2015): 246.

\bibitem{PaperELadder4}
Sepulveda-Falla, Diego, et al. "Familial Alzheimer’s disease–associated presenilin-1 alters cerebellar activity and calcium homeostasis." The Journal of clinical investigation 124.4 (2014): 1552-1567.

\bibitem{PaperELadder5}
Veloz, María Fernanda Vinueza, et al. "Cerebellar control of gait and interlimb coordination." Brain Structure and Function 220.6 (2015): 3513-3536.

\bibitem{PaperCatWalk}
\url{https://goo.gl/ippZrg}

\bibitem{PaperCatWalk1}
Vidal, Pia M., et al. "Delayed decompression exacerbates ischemia-reperfusion injury in cervical compressive myelopathy." JCI insight 2.11 (2017).

\bibitem{PaperCatWalk2}
Tatenhorst, Lars, et al. "Fasudil attenuates aggregation of α-synuclein in models of Parkinson’s disease." Acta neuropathologica communications 4.1 (2016): 39.

\bibitem{PaperCatWalk3}
Zhou, Ming, et al. "Gait analysis in three different 6-hydroxydopamine rat models of Parkinson's disease." Neuroscience letters 584 (2015): 184-189.

\bibitem{PaperCatWalk4}
Chen, Ying-Ju, et al. "Detection of subtle neurological alterations by the Catwalk XT gait analysis system." Journal of neuroengineering and rehabilitation 11.1 (2014): 62.

\bibitem{PaperCatWalk5}
Hou, Jiamei, et al. "Effect of combined treadmill training and magnetic stimulation on spasticity and gait impairments after cervical spinal cord injury." Journal of neurotrauma 31.12 (2014): 1088-1106.


\bibitem{PaperKain}
Kain, Jamey, et al. "Leg-tracking and automated behavioural classification in Drosophila." Nature communications 4 (2013): 1910.

\bibitem{PaperRoy}
Roy, Snigdha, et al. "High-precision, three-dimensional tracking of mouse whisker movements with optical motion capture technology." Frontiers in behavioral neuroscience 5 (2011): 27.

\bibitem{PaperScott}
Tashman, Scott, and William Anderst. "In-vivo measurement of dynamic joint motion using high speed biplane radiography and CT: application to canine ACL deficiency." Journal of biomechanical engineering 125.2 (2003): 238-245.

\bibitem{PaperHarvey}
A. Harvey, Roberto Bermejo, H. Philip Zeigler, Michael. "Discriminative whisking in the head-fixed rat: optoelectronic monitoring during tactile detection and discrimination tasks." Somatosensory \& motor research 18.3 (2001): 211-222.

\bibitem{PaperBemejo}
Bermejo, R., D. Houben, and H. Philip Zeigler. "Optoelectronic monitoring of individual whisker movements in rats." Journal of neuroscience methods 83.2 (1998): 89-96.


\bibitem{PaperKyme}
Kyme, Andre, et al. "Tracking and characterizing the head motion of unanaesthetized rats in positron emission tomography." Journal of The Royal Society Interface 9.76 (2012): 3094-3107.

\bibitem{PaperKymeMarker}
Kyme, A., Zhou, V., Meikle, S. and Fulton, R. 2008 Real-time 3D motion tracking for small animal brain PET. Phys. Med. Biol. 53, 2651–2666.

\bibitem{PaperKymeSynch}
Kyme, Andre Z., et al. "Optimised motion tracking for positron emission tomography studies of brain function in awake rats." PLoS One 6.7 (2011): e21727.

\bibitem{PaperPasquet}
Pasquet, Matthieu O., et al. "Wireless inertial measurement of head kinematics in freely-moving rats." Scientific reports 6 (2016): 35689.

\bibitem{PaperKnutsen}
Knutsen, Per Magne, Dori Derdikman, and Ehud Ahissar. "Tracking whisker and head movements in unrestrained behaving rodents." Journal of neurophysiology 93.4 (2005): 2294-2301.

\bibitem{PaperGravel}
Gravel, Pierre, et al. "A semi-automated software tool to study treadmill locomotion in the rat: from experiment videos to statistical gait analysis." Journal of neuroscience methods 190.2 (2010): 279-288.

\bibitem{PaperBootstrap}
Lenhoff, Mark W., et al. "Bootstrap prediction and confidence bands: a superior statistical method for analysis of gait data." Gait \& Posture 9.1 (1999): 10-17.

\bibitem{PaperAkihiro}
Nakamura, Akihiro, et al. "Low-cost three-dimensional gait analysis system for mice with an infrared depth sensor." Neuroscience research 100 (2015): 55-62.

\bibitem{PaperAGEX}
Plagemann, C., Ganapathi, V., Koller, D., Thrun, S.,2010. Real-time identification and
localization of body parts from depth images. In: 2010 IEEE International Conference
on Robotics and Automation (ICRA). IEEE, pp. 3108–3113.

\bibitem{PaperMendes}
Mendes, César S., et al. "Quantification of gait parameters in freely walking rodents." BMC biology 13.1 (2015): 50.

\bibitem{MouseWalker}
MouseWalker. http://biooptics.markalab.org/MouseWalker/.
doi:10.5281/zenodo.18233.


\bibitem{PaperWang}
Wang, Zheyuan, S. Abdollah Mirbozorgi, and Maysam Ghovanloo. "Towards a kinect-based behavior recognition and analysis system for small animals." Biomedical Circuits and Systems Conference (BioCAS), 2015 IEEE. IEEE, 2015.

\bibitem{PaperMoteiro}
Monteiro, Joao P., et al. "A depth-map approach for automatic mice behavior recognition." Image Processing (ICIP), 2014 IEEE International Conference on. IEEE, 2014.

\bibitem{PaperVoigts}
Voigts, Jakob, Bert Sakmann, and Tansu Celikel. "Unsupervised whisker tracking in unrestrained behaving animals." Journal of neurophysiology 100.1 (2008): 504-515.

\bibitem{PaperPetrou}
Petrou, Georgios, and Barbara Webb. "Detailed tracking of body and leg movements of a freely walking female cricket during phonotaxis." Journal of neuroscience methods 203.1 (2012): 56-68.

\bibitem{PaperXu}
Xu, Qi, et al. "A video tracking system for limb motion measurement in small animals." Optoelectronics and Image Processing (ICOIP), 2010 International Conference on. Vol. 1. IEEE, 2010.

\bibitem{PaperJohn}
Bender, John A., Elaine M. Simpson, and Roy E. Ritzmann. "Computer-assisted 3D kinematic analysis of all leg joints in walking insects." PloS one 5.10 (2010): e13617.

\bibitem{PaperHwang}
Hwang, Sangchul, and Young Choi. "Tracking the joints of arthropod legs using multiple images and inverse kinematics." International Journal of Precision Engineering and Manufacturing 16.4 (2015): 669-675.

\bibitem{PaperFABRIK}
Aristidou, Andreas, and Joan Lasenby. "FABRIK: a fast, iterative solver for the inverse kinematics problem." Graphical Models 73.5 (2011): 243-260.

\bibitem{PaperGyory}
Gyory, G., et al. "An algorithm for automatic tracking of rat whiskers." Proc Int Workshop on Visual observation and Analysis of Animal and Insect Behavior (VAIB), Istanbul, in conjunction with ICPR. Vol. 2010. 2010.

\bibitem{PaperHamers}
Hamers, Frank PT, et al. "Automated quantitative gait analysis during overground locomotion in the rat: its application to spinal cord contusion and transection injuries." Journal of neurotrauma 18.2 (2001): 187-201.

\bibitem{PaperDaSilva}
Da Silva Aragão, Raquel, et al. "Automatic system for analysis of locomotor activity in rodents-A reproducibility study." Journal of neuroscience methods 195.2 (2011): 216-221.

\bibitem{PaperLeroy}
Leroy, Toon, et al. "Automatic analysis of altered gait in arylsulphatase A-deficient mice in the open field." Behavior research methods 41.3 (2009): 787-794.

\bibitem{PaperDankert}
Dankert, Heiko, et al. "Automated monitoring and analysis of social behavior in Drosophila." Nature methods 6.4 (2009): 297.

\bibitem{PaperGMM}
Bishop, CM. Pattern Recognition and Machine Learning. Springer; New York: 2007. p. 738

\bibitem{PaperOtsu}
Otsu N. A threshold selection method from gray level histograms. IEEE Trans Systems, Man and
Cybernetics 1979;9:62–66.

\bibitem{PaperNathan}
Clack, Nathan G., et al. "Automated tracking of whiskers in videos of head fixed rodents." PLoS computational biology 8.7 (2012): e1002591.

\bibitem{PaperKim}
Kim, Hyun June, et al. "Semi-Automated Tracking of Vibrissal Movements in Free-Moving Rodents Captured by High-Speed Videos." World Academy of Science, Engineering and Technology, International Journal of Biological, Biomolecular, Agricultural, Food and Biotechnological Engineering 9.5 (2015): 565-569.

\bibitem{PaperPalmer9}
Palmér, Tobias, et al. "Rat Paw Tracking for Detailed Motion Analysis." Visual observation and analysis of Vertebrate And Insect Behavior 2014. 2014.

\bibitem{PaperPalmer}
Palmer, Tobias, et al. "A system for automated tracking of motor components in neurophysiological research." Journal of neuroscience methods 205.2 (2012): 334-344.


\bibitem{PaperOurs}
Giovannucci, A., et al. "Automated gesture tracking in head-fixed mice." Journal of neuroscience methods (2017).

\bibitem{PaperAGATHA}
Kloefkorn, Heidi E., et al. "Automated Gait Analysis Through Hues and Areas (AGATHA): a method to characterize the spatiotemporal pattern of rat gait." Annals of biomedical engineering 45.3 (2017): 711-725.


















\end{thebibliography}
\end{document}